\documentclass[preprint,12pt]{elsarticle}

\usepackage{amssymb}
\usepackage{amsmath}

\usepackage[utf8]{inputenc} 
\usepackage[T1]{fontenc}    

\usepackage{amsmath}
\usepackage{amsfonts}       
\usepackage{nicefrac}       

\usepackage{graphicx}
\usepackage{booktabs}       
\usepackage{multirow}
\usepackage{colortbl}
\usepackage[table]{xcolor}  

\usepackage{microtype}      
\usepackage{wrapfig}

\usepackage{hyperref}       
\usepackage{url}            

\usepackage[linesnumbered,ruled,vlined]{algorithm2e} 
\usepackage{orcidlink}      



\journal{Nuclear Physics B}

\begin{document}

\begin{frontmatter}

\title{Open-World Test-Time Adaptation with Hierarchical Feature Aggregation and Attention Affine}

\author[label1]{Ziqiong Liu\corref{cofirst}}
\author[label1]{Yushun Tang\corref{cofirst}}
\author[label1,label2]{Junyang Ji}
\author[label1]{Zhihai He\corref{corr}}

\cortext[cofirst]{Equal contributions.}
\cortext[corr]{Corresponding author.}

\affiliation[label1]{%
  organization={Southern University of Science and Technology},
  city={Shenzhen},
  country={China}}
\affiliation[label2]{%
  organization={Shenzhen International Graduate School, Tsinghua University},
  city={Shenzhen},
  country={China}}

\begin{abstract}
Test-time adaptation (TTA) refers to adjusting the model during the testing phase to cope with changes in sample distribution and enhance the model's adaptability to new environments. In real-world scenarios, models often encounter samples from unseen (out-of-distribution, OOD) categories. Misclassifying these as known (in-distribution, ID) classes not only degrades predictive accuracy but can also impair the adaptation process, leading to further errors on subsequent ID samples. Many existing TTA methods suffer substantial performance drops under such conditions. To address this challenge, we propose a Hierarchical Ladder Network that extracts OOD features from class tokens aggregated across all Transformer layers. OOD detection performance is enhanced by combining the original model prediction with the output of the Hierarchical Ladder Network (HLN) via weighted probability fusion. To improve robustness under domain shift, we further introduce an Attention Affine Network (AAN) that adaptively refines the self-attention mechanism conditioned on the token information to better adapt to domain drift, thereby improving the classification performance of the model on datasets with domain shift. Additionally, a weighted entropy mechanism is employed to dynamically suppress the influence of low-confidence samples during adaptation. Experimental results on benchmark datasets show that our method significantly improves the performance on the most widely used classification datasets. 
\end{abstract}


\end{frontmatter}


\section{Introduction}
In real-world applications, machine learning models often encounter samples whose distributions differ significantly from the training data. Such discrepancies may arise from novel categories, sensor noise, or domain shifts~\cite{hendrycks2016baseline, liang2017enhancing, liu2020energy, zhang2023openood, tang2023neuro, yang2024generalized}, leading to a severe degradation of predictive performance in dynamic environments~\cite{recht2019imagenet, wang2021tent, li2023robustness}. 
To mitigate this issue, \textit{Test-Time Adaptation} (TTA) has been proposed. The core idea of TTA is to dynamically update the model using unlabeled target-domain samples encountered during inference, thereby improving adaptability to new environments and distribution shifts without relying on source-domain data or target-domain labels~\cite{wang2021tent, wang2022generalizing}. Compared to conventional domain adaptation methods, TTA is more flexible and practical, making it particularly suitable for real-world applications where annotation is costly or data privacy is restricted~\cite{chakrabarty2023santa, wang2022continual, zhou2023ods, tang2024domain}.

Most existing TTA studies primarily focus on addressing distributional shifts caused by sensor noise or domain corruption. However, in open-world testing scenarios, models may encounter samples from unseen categories or distributions not covered by the training data~\cite{lee2023towards, zhou2023ods}. If these \textit{out-of-distribution} (OOD) samples are incorrectly identified as \textit{in-distribution} (ID) and used for adaptation, they can introduce noisy gradients that propagate through subsequent updates, ultimately leading to performance collapse~\cite{li2023robustness, yu2024stamp, gao2024unified}. Such misadaptation can be especially hazardous in safety-critical domains like autonomous driving and medical diagnosis~\cite{amodei2016concrete, hendrycks2019using}.

To address this challenge, several open-world TTA frameworks have been proposed. 
OSTTA~\cite{lee2023towards} introduces \textit{open-set test-time adaptation} by comparing prediction confidence before and after adaptation, filtering out samples with decreased confidence to avoid noisy updates. 
Despite its simplicity and effectiveness, OSTTA heavily depends on confidence-based selection and lacks explicit OOD identification, limiting its robustness in complex or long-term adaptation scenarios. 
OWTTT~\cite{li2023robustness} adopts a prototype-based clustering strategy with distribution alignment regularization but relies strongly on source-domain data for dynamic prototype expansion. 
STAMP~\cite{yu2024stamp} dynamically updates a class-balanced memory bank with low-entropy, label-consistent samples; however, repeated augmentations and forward inferences for each sample, along with memory maintenance, introduce considerable computational overhead, reducing efficiency in real-time or resource-limited settings.

\begin{figure}[htbp]
    \centering
    \includegraphics[width=1.0\linewidth]{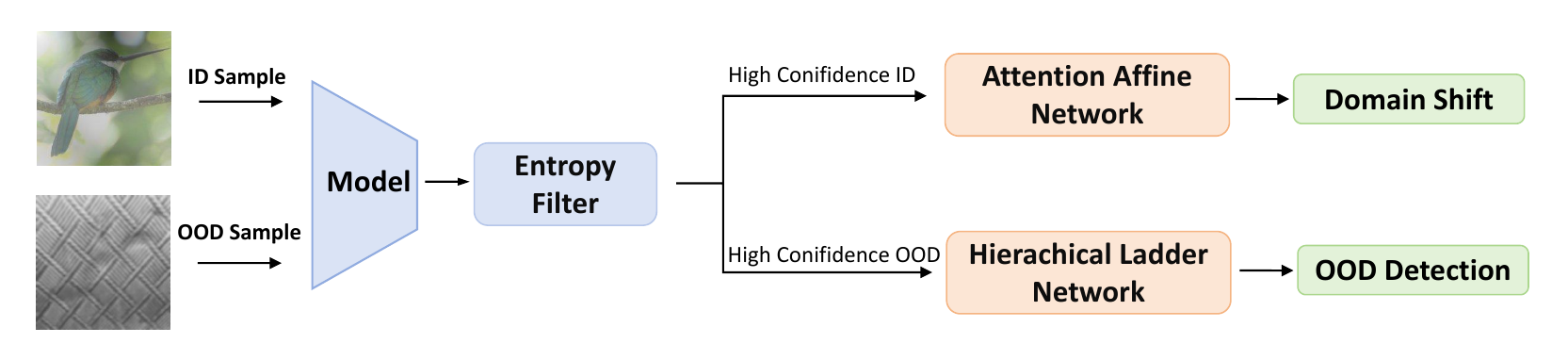}
    \caption{Diagram of our method. An entropy-based filter is applied to identify high-confidence ID and OOD samples. The ID samples are used to update the Attention Affine Network for domain shift adaptation, while the OOD samples are used to update the Hierarchical Ladder Network for improved OOD detection.}
    \label{diagram}
\end{figure}

To further tackle the aforementioned limitations, we propose a novel test-time adaptation (TTA) framework that is explicitly designed to enhance robustness against out-of-distribution (OOD) samples under open-world scenarios.
Our approach incorporates a \textbf{Hierarchical Ladder Network (HLN)} that extracts OOD-relevant features from class tokens aggregated across all layers of the Transformer. Unlike prior methods that rely on shallow statistics or isolated representations, our hierarchical design captures both low-level and high-level semantic discrepancies, enabling more accurate OOD detection. We further refine predictions by fusing outputs from the base model and the HLN using a weighted probability fusion strategy, leading to more calibrated and reliable decisions under distribution shift.

Additionally, we enhance the self-attention mechanism within the Transformer architecture by introducing an \textbf{Attention Affine Network (AAN)}, which leverages token-level information to dynamically adjust attention weights. This allows the model to focus more effectively on domain-relevant features, improving its adaptability to domain drift without compromising performance on well-aligned samples. To stabilize the adaptation process, we also employ a \textbf{weighted entropy mechanism} that downweights low-confidence samples during test-time updates. This not only reduces the adverse impact of uncertain predictions but also ensures robust and consistent adaptation across diverse domain shifts.
We evaluate our method on several benchmark datasets, and experimental results demonstrate its effectiveness. Our framework significantly improves classification performance and consistently outperforms existing TTA methods across various domain shift scenarios.

\section{Related Work and Major Contributions}
\label{related_work}
This work is related to existing research on test-time adaptation and OOD detection. In addition, this section highlights our key contributions in the context of prior work.

\subsection{Test-time Adaptation}
In scenarios where models must adapt to new target domains without access to source data, Test-Time Adaptation (TTA) has emerged as an effective approach for rapid model adjustment with minimal inference overhead \cite{wang2021tent}. Unlike traditional domain adaptation \cite{gebru2017fine, wang2018deep} and domain generalization methods \cite{tong2023distribution, wang2022generalizing}, TTA requires no labeled target data or source samples, making it well-suited for real-world applications with scarce annotations or privacy constraints \cite{ding2023maps, ding2023proxymix, sheng2023adaptguard, tang2024dual, tang2024learning}. Existing TTA methods primarily fall into two categories: Batch Normalization (BN) calibration, which adapts models by updating test batch statistics and combining source and target statistics for robustness \cite{nado2020evaluating, schneider2020improving, yuan2023robust}, and self-training approaches that leverage unsupervised objectives like entropy minimization and contrastive learning \cite{wang2021tent, chen2022contrastive} but can suffer from noisy pseudo-labels. To mitigate this, sample filtering based on entropy has been proposed \cite{niu2022efficient, niu2023towards}. More recent work extends TTA to dynamic domains, label shifts, and temporal changes using specialized optimization strategies \cite{chakrabarty2023santa, wang2022continual, zhou2023ods, zhao2023pitfalls}.
Recent studies such as OWTTT \cite{li2023robustness} and OSTTA \cite{lee2023towards} have investigated test-time scenarios where unknown categories may be present in the data. While both works aim to enhance model generalization in open-world settings, they define distinct problem formulations: OWTTT \cite{li2023robustness} addresses the presence of entirely unknown OOD samples in the test set, whereas OSTTA \cite{lee2023towards} considers OOD samples that share similar distributional shifts with ID samples, such as noise or corruption. To better reflect the complexity of real-world conditions, our experiments are designed to encompass both scenarios.
To address these challenges, a variety of strategies have been proposed. Some works adopt self-training methods based on prototype category expansion for open-world adaptation \cite{li2023robustness}, while others incorporate contrastive learning to improve the separability between ID and OOD representations \cite{su2024open}. However, such approaches typically rely on access to source domain data as auxiliary support. An alternative line of work focuses on entropy-based optimization. For instance, OSTTA \cite{lee2023towards}, UniEnt \cite{gao2024unified}, and AEO \cite{dong2025towards} employ entropy minimization or regularization to distinguish OOD from ID samples. Additionally, STAMP \cite{yu2024stamp} introduces a self-weighted entropy optimization framework combined with a stable memory replay mechanism to enhance robustness.

\subsection{OOD Detection}

Out-of-distribution (OOD) detection is essential for ensuring the reliability and safety of deep learning models deployed in real-world applications \cite{yang2024generalized}. Although these models typically perform well on in-distribution data, they often yield overconfident and unreliable predictions when presented with inputs that deviate from the training distribution, which can result in critical failures in high-risk settings \cite{filos2020can, li2023robustness}. To mitigate this issue, numerous approaches have been proposed to detect OOD samples by analyzing model outputs without modifying the model architecture or training procedure. Common methods include maximum softmax probability (MSP) \cite{hendrycks2016baseline} and its variants such as ODIN \cite{hsu2020generalized, liang2017enhancing}, energy-based techniques \cite{lin2021mood}, and Mahalanobis distance-based scoring \cite{lee2018simple}. Furthermore, recent work demonstrates that incorporating additional auxiliary or outlier samples beyond the training data can enhance the model’s ability to characterize the broader data distribution, thereby improving OOD detection performance and robustness against unseen inputs \cite{park2023powerfulness, zhang2023mixture}.

In recent years, OOD detection research has evolved into a more comprehensive domain. Early approaches employed generative models to detect OOD samples based on reconstruction errors \cite{graham2023denoising}. Follow-up work demonstrated that differences between original and reconstructed images \cite{liu2023unsupervised}, as well as dynamic features of the diffusion process trajectories—such as path change rate and curvature—can effectively identify OOD data \cite{heng2024out, gao2023diffguard}. Beyond visual data, multimodal fusion techniques have also improved detection performance. For instance, MCM \cite{ming2022delving} jointly models images and text to detect OOD samples, and introducing negative prompts has been shown to enhance detection effectiveness across different scenarios \cite{li2024learning, nie2024out, wang2023clipn}.

\subsection{Major Contributions}
\label{sec:contributions}
Compared to existing work, the \textbf{major contributions} of this work can be summarized as: 
(1) We propose a novel OOD detection method that leverages Class token information to extract discriminative features for unknown categories, effectively reducing misclassification under open-set conditions. In addition, the method maintains high ID classification accuracy by preventing excessive adaptation to uncertain samples, thus achieving a better balance between robustness and stability during test-time adaptation. 
(2) We introduce an improved attention mechanism based on token information to enhance the model's adaptability to domain shift. Specifically, we design a QKV affine adaptation strategy that dynamically calibrates the Query, Key, and Value projections during test time, enabling better alignment of internal representations with the target domain and improving adaptation performance. 
(3) We conduct extensive experiments on several widely-used domain shift benchmarks, demonstrating that our method consistently maintains high classification accuracy on in-distribution (ID) samples, even under the interference of OOD inputs. This performance gain is primarily attributed to our effective integration of OOD-aware mechanisms into the test-time adaptation framework.

\section{Method}
\label{headings}

In this section, we present our method of Hierarchical Feature Aggregation and Attention Affine for open-world test-time adaptation.

\subsection{Problem Formulation and Method Overview}
Suppose that a model $\mathcal{M}=f_{\theta_s}(y|X_s)$ with parameters $\theta_s$ has been successfully trained on the source data $\{X_s\}$ with labels $\{Y_s\}$. 
During open-world test-time adaptation, we are given the target data $\{X_t\}$ with unknown labels $\{Y_t\}$, where $\{Y_s\} \subseteq \{Y_t\}$. Our goal is to adapt the trained model to recognize the in-distribution samples and identify out-of-distribution instances to reject recognition, in an unsupervised manner during testing. Given a sequence of input sample batches $\{\mathbf{B}_1, \mathbf{B}_2, ..., \mathbf{B}_T\}$ from $\{X_t\}$, the $t$-th adaptation of the network model can only rely on the $t$-th batch of test samples $\mathbf{B}_t$.

\begin{figure}[htbp]
    \centering
    \includegraphics[width=1.0\linewidth]{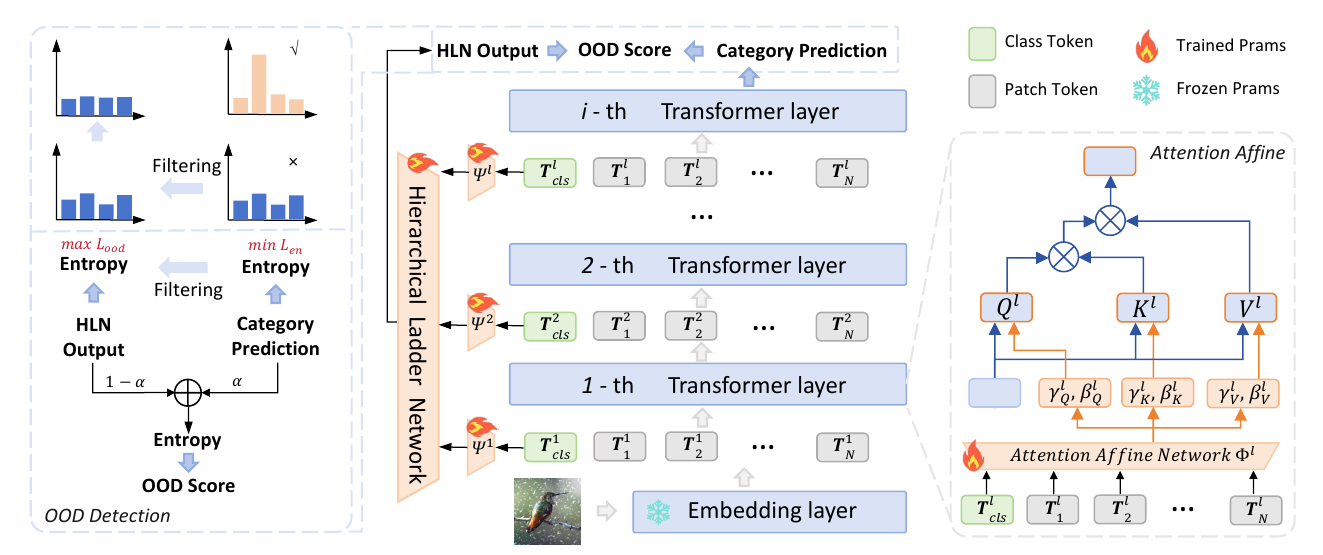}
    \caption{Overview of our proposed Hierarchical Feature Aggregation and Attention Affine.}
    \label{overview}
\end{figure}

Our method aims to dynamically adjust the Vision Transformer model during inference to mitigate performance degradation caused by domain shifts and OOD samples. The method overview is shown in Figure \ref{overview}. In this work, we propose a Hierarchical Ladder Network that extracts out-of-distribution (OOD) features from class tokens aggregated across all Transformer layers. OOD detection performance is enhanced by combining the original model prediction with the output of the Hierarchical Ladder Network (HLN) via weighted probability fusion. To improve robustness under domain shift, we further introduce an Attention Affine Network (AAN) that adaptively refines the self-attention mechanism conditioned on the token information to better adapt to domain drift, thereby improving the classification performance of the model on datasets with domain shift. Additionally, a weighted entropy mechanism is employed to dynamically suppress the influence of low-confidence samples during adaptation.
In the following sections, we will further explain the proposed method in more detail.

\subsection{Learning the Hierarchical Ladder Network for OOD Detection}
\begin{figure}[htbp]
    \centering
    \includegraphics[width=1.0\linewidth]{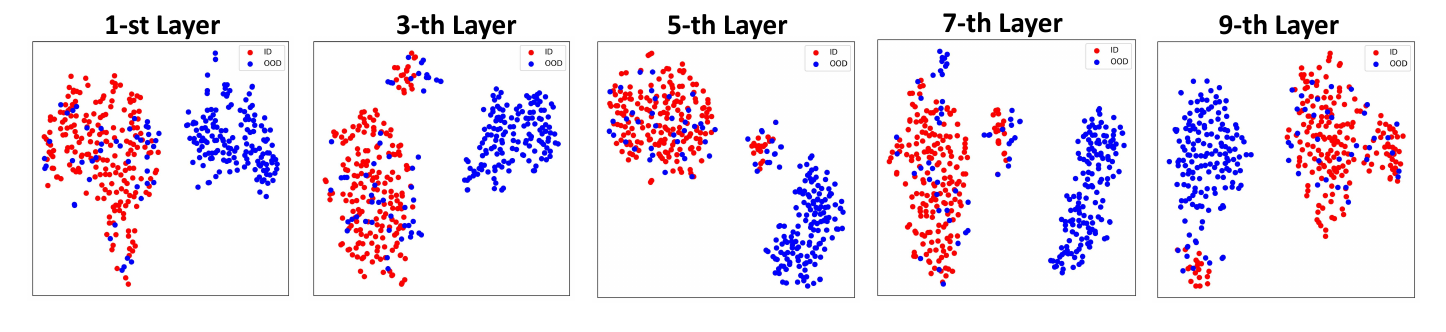}
    \caption{T-SNE visualization of the Class Tokens from different layers.}
    \label{fig:tsne}
\end{figure}

Inspired by \cite{yu2024stamp, gao2024unified}, we also observe that during test-time adaptation, samples with lower entropy are more likely to belong to in-distribution (ID) classes, as they typically exhibit a sharp distribution of predicted logits. Unlike methods such as \cite{gao2024unified} and \cite{li2023robustness}, which leverage class prototypes from the source domain as reference, we follow the setting in \cite{yu2024stamp} and perform entropy optimization based on reliable samples. Additionally, as shown in the t-SNE visualization results in Figure~\ref{fig:tsne}, we observe that the separable information between OOD and ID samples is not only present in the last-layer class token, but also exists across class tokens from intermediate layers. Motivated by this observation, we incorporate the Hierarchical Ladder Network for high-confidence OOD samples to leverage multi-layer class token representations. This design enhances the separability between OOD and ID samples while maintaining classification accuracy.

\textbf{Hierarchical Ladder Network.} 
Before introducing the Hierarchical Ladder Network(HLN), we define an out-of-distribution (OOD) feature extractor \( \Psi^l \), which operates on each layer of the transformer to extract information from class token. Specifically, at the \( l \)-th layer, let the class token be denoted as \( \mathbf{c}_{\text{cls}}^{(l)} \in \mathbb{R}^{d} \), where \( d \) is the feature dimension. The OOD feature extractor takes \( \mathbf{c}_{\text{cls}}^{(l)} \) as input and generates an \textit{OOD token} \( \mathbf{o}^{(l)} \) as follows:
\begin{equation}
\mathbf{o}^{(l)} = \Psi^l\left( \mathbf{c}_{\text{cls}}^{(l)} \right), \quad l = 1, 2, \dots, L.
\end{equation}
where $L$ is the total number of transformer layers. Notably, $\Psi$ is shared across all layers. 
The OOD token $\mathbf{o}^{(l)}$ is designed to carry richer out-of-distribution cues than the original class token and is used in subsequent components of the framework. The OOD token $\mathbf{o}^{(l)}$ from each layer is then forwarded to the Hierarchical Ladder Network (HLN) to generate a comprehensive OOD information token. Formally, the final OOD token is obtained as:
\begin{equation}
\mathbf{o}^{\text{hln}} = \mathrm{HLN}(\{ \mathbf{o}^{(l)} \}_{l=1}^L),
\end{equation}
where $\mathrm{HLN}(\cdot)$ denotes the aggregation function implemented by the Hierarchical Ladder Network. The final prediction is produced by feeding $\mathbf{o}^{\text{hln}}$ into a classifier $\mathcal{C}$, which is shared with the original class token, as follows:
\begin{equation}
\mathbf{p}^{\text{OOD}} = \mathrm{softmax}(\mathcal{C}(\mathbf{o}^{\text{hln}})),
\end{equation}
where $\mathbf{p}^{\text{OOD}}$ denotes the resulting probability distribution. Based on this distribution, the entropy of the OOD token for sample $i$ is computed as \( H_i^{\text{OOD}} = - \sum_{c=1}^{C} p_{i,c}^{\text{OOD}} \log p_{i,c}^{\text{OOD}} \), where \( C \) is the total number of classes and \( p_{i,c}^{\text{OOD}} \) is the predicted probability for class \( c \).

\textbf{Loss for OOD.} To encourage the model to better distinguish between in-distribution and out-of-distribution samples based on both the original model category prediction and the dedicated OOD branch, we introduce an OOD-specific loss. Specifically, we define a binary mask \( \mathbf{m} \in \{0,1\}^N \) using a predefined entropy threshold \( \mathcal{H}_{\text{thr}}^{\text{OOD}} \):
\begin{equation}
m_i = \mathbb{I}(\mathcal{H}_i > \mathcal{H}_{\text{thr}}^{\text{OOD}}),
\end{equation}

where \( \mathbb{I}(\cdot) \) denotes the indicator function, \( \mathcal{H}_i = - \sum_{c=1}^{C} p_{i, c} \log p_{i, c} \) represents the entropy of sample \( i \) in \( B_t \), and \( N \) is the number of samples in the current batch. 
As illustrated in the left part of Figure~\ref{overview}, the entropy \( \mathcal{H}_i \) is computed based on the category prediction of sample \( i \). This filtering process facilitates subsequent operations on the HLN outputs for OOD samples by selecting only those with relatively uniform (i.e., high-entropy) predictions. The OOD loss is then defined as the average negative entropy over the masked samples:
\[
\mathcal{L}_{\text{OOD}} = -\frac{1}{\sum_{i=1}^N \mathbf{m}_i} \sum_{i=1}^{N} \mathbf{m}_i \cdot \mathcal{H}^{\text{OOD}}_i,
\]
where \( \mathcal{H}^{\text{OOD}}_i \) is the entropy computed by the OOD branch. The motivation of this part is to encourage the outputs of OOD samples after the HLN network to become more uniform, which is beneficial for using OOD scores to detect such samples.

\textbf{OOD Score.} After the above optimization steps, we obtain the final OOD score by computing a weighted combination of the in-distribution and OOD predictions from the classifier \( \mathcal{C} \):
\begin{equation}
\mathbf{p}^{\text{final}} = \alpha \cdot \mathrm{softmax}(\mathcal{C}({\mathbf{c}_\text{cls}}^{(L)})) + (1 - \alpha) \cdot \mathrm{softmax}(\mathcal{C}(\mathbf{o}^{\text{hln}})),
\end{equation}
where \( \alpha \in [0, 1] \) is a hyperparameter controlling the balance between in-distribution and OOD predictions, and its sensitivity is analyzed in section \ref{ood_hyper}. Here ${\mathbf{c}_\text{cls}}^{(L)}$ denotes the class token at the last layer. The final OOD score is defined as the entropy of the combined output:
\begin{equation}
\mathcal{H}_{\text{i}}^{\text{final}} = - \sum_{c=1}^{C} p_{i, c}^{\text{final}} \log p_{i, c}^{\text{final}},
\end{equation}
where $p_{i, c}^{\text{final}}$ is the $i$-th element of $\mathbf{p}^{\text{final}}$.

\subsection{Learning the Attention Affine Network for Domain Shift Correction}

The self-attention mechanism is a core component of Transformer models, leveraging Query, Key, and Value (QKV) projections to compute attention weights and produce contextualized representations. However, these projections are highly sensitive to domain shifts, often resulting in degraded performance when applied to novel target domains. To address this issue, we propose the Attention Affine Network (AAN), which dynamically adjusts the QKV projections conditioned on intermediate features at each Transformer layer. This adaptation enables more robust attention computation under domain drift and enhances generalization to unseen environments.
For each attention head, AAN generates scaling/shift parameters based on patch token embeddings $E$:
\begin{equation}
    [\gamma^l_Q, \beta^l_Q; \gamma^l_K, \beta^l_K; \gamma^l_V, \beta^l_V] = \Phi^l(E).
\end{equation}

Formally, the affined Query, Key, and Value components are computed as:
\begin{equation}
\begin{cases}
    Q'^l = \gamma^l_Q \cdot Q^l + \beta^l_Q,\\
    K'^l = \gamma^l_K \cdot K^l + \beta^l_K,\\
    V'^l = \gamma^l_V \cdot V^l + \beta^l_V,\\
\end{cases}
\end{equation}
where $\gamma^l_Q, \gamma^l_K, \gamma^l_V$ are the scale factors and $\beta^l_Q, \beta^l_K, \beta^l_V$ are the shift factors for the Query, Key, and Value components in the $l$-th layer, respectively.

\textbf{Loss for Domain Shift} 
Since all tokens contain information and undergo similar corruptions, they may tend to capture domain shift information when assigned more probable features. To encourage similarity among patch tokens, we define the following patch-wise similarity loss:
\begin{equation}
    \mathcal{L}_{\text{sim}} = - \frac{1}{N_{\text{patch}}} \sum_{i=1}^{N_{\text{patch}}} \sum_{j \ne i}^{N_{\text{patch}}} \frac{\mathbf{x}_i^\top \mathbf{x}_j}{\|\mathbf{x}_i\| \cdot \|\mathbf{x}_j\|},
\end{equation}
where \( \mathbf{x}_i \) and \( \mathbf{x}_j \) are the patch tokens, and \( N_{\text{patch}} \) is the number of patch tokens. This loss encourages similarity by maximizing the cosine similarity between patch tokens, helping to improve the model's adaptation to domain shift.

\subsection{Test-time Adaptation for Open-World}
\label{subsec:test-time adaptation for open-world}

In the closed-set Test-Time Adaptation (TTA) setting, a widely adopted strategy is to update the model by minimizing the entropy of the predictive distribution, as proposed in TENT \cite{wang2021tent}. Given a mini-batch of N samples \( B_t \) at timestamp \( t \), the entropy loss is defined as:
\begin{equation}
\mathcal{L}_{\text{ent}}(B_t) = - \frac{1}{N} \sum_{i=1}^N \mathcal{H}_i,
\end{equation}
where \( p(x) = f_{\theta_t}(x) \) denotes the softmax output of the model with parameters \( \theta_t \) and the correspond predicted entropy is \( \mathcal{H}_i = - \sum_{c=1}^{C} p_{i, c} \log p_{i, c}\) for sample i in \( B_t \). 

Following the entropy optimization strategy proposed in~\cite{yu2024stamp}, we adopt a self-weighted entropy optimization approach. For a batch of $N$ samples, the entropy loss is defined as:
\begin{equation}
\mathcal{L}_{\text{entropy}} = \sum_{i=1}^N \left( \frac{1 / e^{\mathcal{H}_i}}{\sum_{j=1}^N 1 / e^{\mathcal{H}_j}} \cdot N \right) \cdot \mathcal{H}_i,
\end{equation}
where $\mathcal{H}_i$ denotes the entropy of the $i$-th sample. The total loss is composed of three terms:
\begin{equation}
\mathcal{L} = \mathcal{L}_{\text{entropy}} + \beta_{1} \mathcal{L}_{\text{OOD}} + \beta_{2} \mathcal{L}_{\text{sim}},
\end{equation}
where $\beta_{1}$ and $\beta_{2}$ hyperparameters. The total loss function encourages confident predictions for in-distribution data and high uncertainty for potential out-of-distribution (OOD) samples, thereby enhancing model robustness in open-world scenarios.

\section{Experiments}
\label{others}

In this section, we evaluate the effectiveness of our proposed method under the open-world test-time adaptation setting across multiple benchmark datasets. The datasets and implementation details are provided in Section~\ref{sec:dataset}, and the performance results are presented in Section~\ref{sec:results}.

\subsection{Datasets and Implementation Details}
\label{sec:dataset}
\textbf{Datasets.} 
In the experiments, we adopt the widely used \textbf{ImageNet-C} benchmark dataset \cite{hendrycks2019benchmarking}, which covers all categories of ImageNet-1k, including 15 types of image degradation, 5 intensity levels, and 50,000 images per level. We also use the common ImageNet-1k OOD datasets, including subsets of \textbf{Texture} \cite{cimpoi2014describing}, \textbf{Places} \cite{zhou2017places}, \textbf{iNaturalist} \cite{van2018inaturalist} and \textbf{SUN} \cite{xiao2010sun}. To evaluate the robustness of the method under the domain drift scenario, we refer to the setting of \cite{yu2024stamp} and construct the corresponding perturbation datasets \textbf{Texture-C} and \textbf{Places-C}. Additionally, we evaluate on \textbf{ImageNet-R} \cite{hendrycks2021many}, which is mainly used to test the generalization ability of the model under style transfer conditions and contains 30,000 images with diverse styles. We also include \textbf{ImageNet-A} \cite{hendrycks2021natural}, a challenging subset of ImageNet-1k consisting of naturally distributed samples that are particularly prone to classification errors. 

We compare our approach with the source model (without adaptation) serving as the baseline. Our method is compared against several state-of-the-art online test-time adaptation (TTA) approaches, including TENT \cite{wang2021tent}, CoTTA \cite{wang2022continual}, SoTTA \cite{gong2023sotta}, SAR \cite{niu2023towards}, OSTTA \cite{lee2023towards}, UniEnt \cite{gao2024unified} and STAMP \cite{yu2024stamp}, with the source model (without adaptation) serving as the baseline.

\begin{table*}[!htbp]
\begin{center}
\caption{Classification Accuracy (\%) for each corruption type in \textbf{ImageNet-C} at the highest severity level (Level 5) under attacks from OOD datasets (Textures, Places, iNaturalist, and SUN). The best result is shown in \textbf{bold}.}
\label{table:ImageNet_c}
\resizebox{\linewidth}{!}
{
\begin{tabular}{l|ccc|ccc|ccc|ccc|ccc}
\toprule
\multirow{2}{*}{Method} & \multicolumn{3}{c|}{Textures} & \multicolumn{3}{c|}{Places} & \multicolumn{3}{c|}{iNaturalist} & \multicolumn{3}{c|}{SUN} & \multicolumn{3}{c}{Avg.} \\
& ACC & AUC & H-score & ACC & AUC & H-score & ACC & AUC & H-score & ACC & AUC & H-score & ACC & AUC & H-score \\
\midrule

Source & 29.9 & 64.9 & 39.6 & 29.9 & 34.9 & 32.1 & 
29.9 & 54.1 & 38.3 & 29.9 & 38.2 & 33.4 & 29.9 & 48.0 & 35.9 \\ 
Tent & 40.9 & 60.7 & 46.9 & 41.5 & 55.2 & 45.2 & 
40.8 & 72.9 & 50.4 & 41.0 & 58.4 & 46.0 & 41.0 & 61.8 & 47.1 \\

CoTTA & 54.5 & 62.5 & 57.8 & 53.1 & 59.9 & 53.6 & 51.7 & 69.7 & 57.4 & 53.6 & 50.1 & 54.0 & 53.2 & 60.5 & 55.7\\
SoTTA & 56.6 & 69.0 & 59.6 & 56.5 & 66.8 & 58.6 & 
59.9 & 80.3 & 68.5 & 56.8 & 68.6 & 59.7 & 57.4 & 71.2 & 61.6 \\ 
SAR & 64.0 & 70.8 & 67.2 & 62.7 & 68.4 & 65.2 & 
58.8 & 49.1 & 52.1 & 60.4 & 67.7 & 63.3 & 61.5 & 64.0 & 62.0 \\

OSTTA & 39.4 & 60.0 & 45.0 & 39.0 & 59.0 & 44.5 & 
39.1 & 58.7 & 44.4 & 39.1 & 57.9 & 44.1 & 39.2 & 58.9 & 43.9 \\
UniEnt & 37.9 & 51.5 & 43.0 & 37.3 & 42.8 & 39.6 & 
38.1 & 66.3 & 47.9 & 37.3 & 44.6 & 40.4 & 37.7 & 51.3 & 42.7 \\

STAMP & 59.8 & 72.6 & 64.1 & 62.8 & 65.4 & 64.0 & 
63.8 & 88.9 & 73.7 & 62.6 & 68.8 & 65.6 & 62.0 & 73.9 & 66.8 \\ 

\rowcolor{gray!20}
\textbf{Ours} & \textbf{65.2} & \textbf{85.1} & \textbf{73.7} & \textbf{63.5} & \textbf{75.3} & \textbf{69.2} & 
\textbf{65.3} & \textbf{98.2} & \textbf{78.2} & 
\textbf{64.6} & \textbf{73.3} & \textbf{68.3} & 
\textbf{64.7} & \textbf{82.9} & \textbf{72.3} \\ 
\rowcolor{gray!20}
& ${\pm0.5}$ & ${\pm0.5}$ & ${\pm0.6}$ & 
${\pm0.6}$ & ${\pm0.6}$ & ${\pm0.5}$ & 
${\pm0.3}$ & ${\pm0.2}$ & ${\pm0.2}$ & 
${\pm0.5}$ & ${\pm1.4}$ & ${\pm0.7}$ & 
${\pm0.5}$ & ${\pm0.7}$ & ${\pm0.5}$ \\ 
\bottomrule
\end{tabular}
}
\end{center}
\end{table*}

\textbf{Implementation Details.}

In our experiments, all methods use the same architecture and pre-trained parameters for fair comparison. We use ViT-B/16 as backbone with a batch size of 32. The Attention Affine Network \( \phi \) has two parts: a Token Feature Extraction Network extracting features from patch tokens, and a QKV Affine Network that combines these with the class token to independently predict shift and scale parameters for Query, Key, and Value. The QKV Affine Network is a linear layer with input \( d \) and output \( 6d \), generating affine transformation weights and biases. The Token Feature Extraction Network and \( \Psi \) are fully connected layers of dimension \( d \), where \( \Psi \), shared across layers, extracts class token information. These features feed into the Hierarchical Ladder Network, a fully connected layer with input \( 12d \) and output \( d \), integrating information from all layers. \\
The learning rates for the Token Feature Extraction Network and the QKV Affine network \( \phi \) are set to $0.2$ and $0.0001$, adding about 4.1M parameters. The OOD detection component adds 0.6M parameters, with learning rates for \( \Psi \) and the Hierarchical Ladder Network at $0.1$ and $0.001$, respectively. Training uses SGD optimizer. Following SAR~\cite{niu2023towards}, two backward passes are done per iteration: one for adapted parameters with regularization, one to update model parameters, simulating test-time adaptation. Results report mean and std over three runs with seeds {2024, 2025, 2026}. All experiments were run on an NVIDIA RTX3090 GPU. The source code will be released.

\subsection{Performance Results}
\label{sec:results}

\begin{table}[!htbp]
\centering
\caption{Classification Accuracy (\%) on \textbf{ImageNet-C} and OOD datasets (Textures-C, Places-C) under the same corruption types at severity level 5. Best results are shown in \textbf{bold}.
}
\label{table:cc}
\resizebox{0.7\linewidth}{!}
{
\begin{tabular}{l|ccc|ccc|ccc}
\toprule
\multirow{2}{*}{Method} & \multicolumn{3}{c|}{Textures-C} & \multicolumn{3}{c|}{Places-C} & \multicolumn{3}{c}{Avg.} \\
 & ACC & AUC & H-score & ACC & AUC & H-score & ACC & AUC & H-score \\
\midrule
Source & 29.9 & 64.9 & 39.6 & 29.9 & 34.9 & 32.1 & 29.9 & 53.9 & 37.0 \\ 
Tent & 40.1 & 67.7 & 48.9 & 39.1 & 62.5 & 46.6 & 39.6 & 65.1 & 47.7 \\ 
CoTTA & 52.9 & 67.9 & 57.9 & 54.7 & 65.6 & 59.4 & 53.8 & 66.7 & 58.6 \\ 
SoTTA & 56.2 & 69.8 & 60.7 & 56.5 & 68.1 & 60.7 & 56.4 & 69.0 & 60.7 \\ 
SAR & 57.6 & 65.3 & 60.9 & 57.7 & 67.6 & 62.0 & 57.6 & 66.4 & 61.4 \\ 

OSTTA & 39.1 & 55.8 & 44.2 & 37.8 & 56.5 & 43.5 & 38.4 & 56.2 & 43.9 \\ 
UniEnt & 38.2 & 69.4 & 48.1 & 37.8 & 67.8 & 47.0 & 38.0 & 68.6 & 47.6 \\ 

STAMP & 59.0 & 79.0 & 65.8 & 63.1 & 76.0 & 68.8 & 61.1 & 77.5 & 67.3 \\ 
\rowcolor{gray!20}
\textbf{Ours} & \textbf{64.9} & \textbf{86.0} & \textbf{73.6} & \textbf{65.1} & \textbf{77.3} & \textbf{70.6} & \textbf{65.0} & \textbf{81.7} & \textbf{72.1} \\ 
\rowcolor{gray!20}
& $\pm$0.1 & $\pm$0.4 & $\pm$0.2 & $\pm$0.6 & $\pm$0.2 & $\pm$0.3 & $\pm$0.4 & $\pm$0.3 & $\pm$0.2 \\
\bottomrule
\end{tabular}
}
\end{table}

\begin{table*}[!htbp]

\begin{center}
\caption{Classification Accuracy (\%) for each corruption in \textbf{ImageNet-R/A} at the highest severity (Level 5). The best result is shown in \textbf{bold}.}
\label{table:Imagenetr}
\resizebox{\linewidth}{!}
{
\begin{tabular}{l|ccc|ccc|ccc|ccc|ccc}
\toprule
\multirow{2}{*}{Source} & \multicolumn{3}{c|}{Textures} & \multicolumn{3}{c|}{Places} & \multicolumn{3}{c|}{iNaturalist} & \multicolumn{3}{c|}{SUN} & \multicolumn{3}{c}{Avg.} \\
& ACC & AUC & H-score & ACC & AUC & H-score & ACC & AUC & H-score & ACC & AUC & H-score & ACC & AUC & H-score \\
\midrule

\midrule
\multicolumn{16}{c}{\textbf{ImageNet-R}}\\
\midrule
Source & 42.7 & 58.2 & 49.2 & 42.7 & 52.5 & 47.1 & 
42.7 & 60.5 & 50.1 & 42.7 & 55.7 & 48.3 & 42.7 & 46.7 & 48.7 \\ 
SAR & 60.8 & 69.3 & 64.8 & 61.5 & 68.8 & 64.9 & 
55.4 & 62.8 & 58.8 & 54.9 & 62.1 & 58.3 & 58.1 & 65.8 & 61.7 \\ 
STAMP & 59.3 & 73.3 & 65.5 & 59.3 & 66.5 & 62.7 & 
59.6 & 73.4 & 65.8 & 59.4 & 71.2 & 64.8 & 59.4 & 71.1 & 64.7 \\ 
 
\rowcolor{gray!20}
\textbf{Ours} & \textbf{63.2} & \textbf{79.8} & \textbf{70.5} & \textbf{63.3} & \textbf{75.2} & \textbf{68.8} & 
\textbf{62.9} & \textbf{84.2} & \textbf{72.0} & 
\textbf{63.0} & \textbf{82.6} & \textbf{71.5} & 
\textbf{63.1} & \textbf{80.5} & \textbf{70.7} \\ 
\rowcolor{gray!20}
& ${\pm0.3}$ & ${\pm0.4}$ & ${\pm0.3}$ & 
${\pm0.6}$ & ${\pm1.6}$ & ${\pm1.0}$ & 
${\pm0.4}$ & ${\pm0.4}$ & ${\pm0.4}$ & 
${\pm0.4}$ & ${\pm0.8}$ & ${\pm0.6}$ & 
${\pm0.4}$ & ${\pm0.8}$ & ${\pm0.6}$ \\ 
\midrule

\multicolumn{16}{c}{\textbf{ImageNet-A}}\\
\midrule
Source & 22.6 & 65.7 & 33.6 & 22.6 & 55.9 & 32.2 & 
22.6 & 68.8 & 34.0 & 22.6 & 63.1 & 33.3 & 42.7 & 46.7 & 48.7 \\ 
SAR & 25.2 & 68.0 & 64.8 & 25.0 & 57.3 & 34.8 & 
25.0 & 71.6 & 37.1 & 24.5 & 62.7 & 35.3 & 24.9 & 64.9 & 36.0 \\ 
STAMP & 31.0 & 69.0 & 42.8 & 30.8 & 59.4 & 43.7 & 
30.8 & \textbf{75.0} & 43.7 & 30.9 & \textbf{67.2} & 42.3 & 30.9 & 67.6 & 42.3 \\ 

\rowcolor{gray!20}
\textbf{Ours} & \textbf{39.2} & \textbf{78.9} & \textbf{52.4} & \textbf{38.9} & \textbf{67.3} & \textbf{49.3} & 
\textbf{38.8} & 70.8 & \textbf{50.1} & 
\textbf{34.8} & 64.1 & \textbf{43.7} & 
\textbf{37.9} & \textbf{70.3} & \textbf{48.9} \\ 
\rowcolor{gray!20}
& ${\pm1.5}$ & ${\pm0.4}$ & ${\pm1.4}$ & 
${\pm0.3}$ & ${\pm0.1}$ & ${\pm0.2}$ & 
${\pm0.4}$ & ${\pm0.4}$ & ${\pm0.4}$ & 
${\pm0.4}$ & ${\pm0.3}$ & ${\pm0.4}$ & 
${\pm1.2}$ & ${\pm0.8}$ & ${\pm1.2}$ \\ 
\bottomrule
\end{tabular}
}
\end{center}
\end{table*}

\textbf{Experimental Results.} 
We systematically evaluate the performance of the proposed method on the ImageNet-C dataset and all results are rounded to one decimal place to ensure the accuracy and comparability of the data. All indicators are based on the experimental results of three groups of different random seeds, and the corresponding average values and standard deviations under the same conditions are shown in Table~\ref{table:ImageNet_c}. We follow the definition of AUC and H-score in \cite{yu2024stamp}. The method is thoroughly evaluated on various OOD datasets under corruption severity level 5 on the ImageNet dataset, encompassing 15 typical corruption types such as noise, blur, weather effects, and digital disturbances. Experimental results demonstrate that in four different OOD datasets (Textures, Places, iNaturalist, and SUN), our proposed method consistently achieves superior and stable performance, with an average classification accuracy of 64.7\%, outperforming the second best method by 2.7\%. Furthermore, the average AUROC improves by 9\%, and the overall balanced accuracy measured by the H score increases by 5.5\%.
We also evaluate our method on datasets following the same protocol as \cite{gao2024unified}, where identical corruption types are applied to both ID and OOD samples to better reflect real-world conditions. As shown in Table~\ref{table:cc}, our method outperforms the second-best approach by 3.9\%, 4.2\%, and 4.8\% in terms of ACC, AUC, and H-score, respectively.

To evaluate the generalization and robustness of the proposed method under real-world distribution shift conditions, we conducted comprehensive experiments on several challenging benchmark datasets. As shown in Table~\ref{table:Imagenetr}, our method achieves an H-score of 70.7\% on the ImageNet-R dataset, outperforming the state-of-the-art methods SAR and STAMP by 9.0\% and 4.5\%, respectively, demonstrating its strong adaptability to test-time scenarios with significant appearance variations. Furthermore, on the more challenging ImageNet-A dataset, our method achieves an H-score of 48.9\%, exceeding the second-best method by 6.6\%.

We further extend our experiments to a larger backbone, ViT-L/16. The results, reported in Table~\ref{table:large}, further verify the robustness and stability of our method across different Transformer backbones and diverse datasets. Detailed hyperparameter settings are provided in the Appendix.

\begin{table*}[!htbp]

\begin{center}
\caption{Classification Accuracy (\%) for each corruption in \textbf{ImageNet-R} with ViT-L/16 at the highest severity (Level 5). The best result is shown in \textbf{bold}.}
\label{table:large}
\resizebox{\linewidth}{!}
{
\begin{tabular}{l|ccc|ccc|ccc|ccc|ccc}
\toprule
\multirow{2}{*}{Source} & \multicolumn{3}{c|}{Textures} & \multicolumn{3}{c|}{Places} & \multicolumn{3}{c|}{iNaturalist} & \multicolumn{3}{c|}{SUN} & \multicolumn{3}{c}{Avg.} \\
& ACC & AUC & H-score & ACC & AUC & H-score & ACC & AUC & H-score & ACC & AUC & H-score & ACC & AUC & H-score \\

\midrule
Source & 61.5 & 73.0 & 66.8 & 61.5 & 68.5 & 64.8 & 
61.5 & 79.5 & 69.4 & 61.5 & 70.4 & 65.6 & 61.5 & 72.8 & 66.6 \\ 
SAR & 62.9 & 74.1 & 68.0 & 62.9 & 69.5 & 66.0 & 
62.9 & 71.4 & 66.9 & 63.1 & 79.5 & 70.3 & 62.9 & 73.6 & 67.8 \\ 
STAMP & 65.4 & 76.5 & 70.5 & 65.5 & 70.5 & 67.9 & 
65.4 & 80.3 & 72.1 & 65.4 & 72.1 & 68.6 & 65.4 & 74.8 & 69.8 \\ 
\rowcolor{gray!20}
\textbf{Ours} & \textbf{66.3} & \textbf{79.4} & \textbf{71.6} & \textbf{66.5} & \textbf{77.4} & \textbf{71.5} & 
\textbf{66.2} & \textbf{84.1} & \textbf{74.1} & 
\textbf{66.0} & \textbf{83.1} & \textbf{73.6} & 
\textbf{66.2} & \textbf{81.0} & \textbf{72.7} \\ 
\rowcolor{gray!20}
& ${\pm0.5}$ & ${\pm0.5}$ & ${\pm0.3}$ & 
${\pm0.4}$ & ${\pm1.1}$ & ${\pm0.8}$ & 
${\pm0.7}$ & ${\pm0.2}$ & ${\pm0.6}$ & 
${\pm0.5}$ & ${\pm1.0}$ & ${\pm0.6}$ & 
${\pm0.5}$ & ${\pm0.7}$ & ${\pm0.6}$ \\ 
\bottomrule
\end{tabular}
}
\end{center}
\end{table*}

\begin{figure}[htbp]
    \centering
    \includegraphics[width=1.0\linewidth]{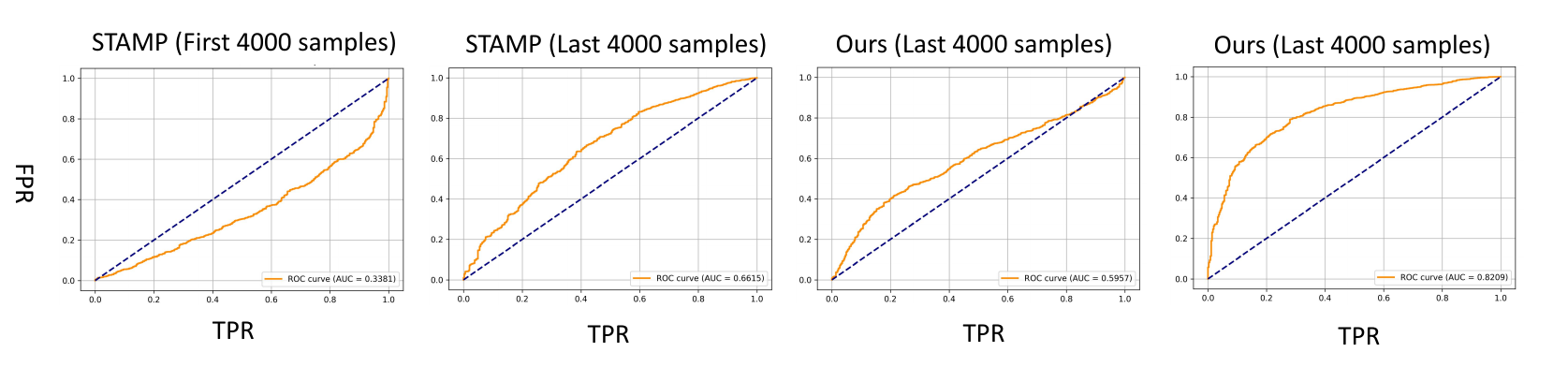}
    \caption{AUROC curve of stamp and our methods with the first 4000 samples and last 4000 samples during adpation.}
    \label{fig:auroc}
\end{figure}

As shown in Figure~\ref{fig:auroc}, we visualize the AUROC curves for the first and last 4000 samples. Our method consistently shows superior discriminative ability across both subsets, which further improves as adaptation proceeds. These results validate the robustness and generalizability of our method in handling diverse OOD samples under complex perturbations. The significant improvements highlight its effectiveness in addressing severe image degradation by leveraging both test-time adaptation and OOD-aware mechanisms.

\tabcolsep=0.3cm
\begin{table}[htbp]
\centering
\small
\caption{Ablation study on ImageNet-C (ID) and Textures (OOD) datasets, evaluating the individual contributions of the Attention Affine module and the Hierarchical Ladder Network.}
\label{table:ablation}
{
\begin{tabular}{cc|cccc}
\toprule
 AAN & HLN & ACC  &AUC  &H-score  \\ \midrule
 - & - & 64.3  &74.3  &68.9  \\
\checkmark & - &65.4  &74.9  & 69.8 \\
-& \checkmark &64.3  &83.9  & 72.7 \\
\checkmark & \checkmark& 65.0  &85.0  & 73.5 \\ 
\bottomrule

\end{tabular}
}
\end{table}

\textbf{Ablation Study.} 
To evaluate the individual contributions of the Attention Affine Network (AAN) and Hierarchical Ladder Network (HLN), we conducted a series of controlled ablation experiments, with results summarized in Table~\ref{table:ablation}. The base model, without either component, achieves 64.3\% ACC, 74.3\% AUC, and a 68.9\% H-score. Introducing the Attention Affine module alone improves all metrics, indicating that affine calibration of Query, Key, and Value within the self-attention module enhances the model’s feature representation under distribution shifts. In contrast, incorporating only the Hierarchical Ladder Network significantly boosts AUC (from 74.3\% to 83.9\%) and H score (from 68.9\% to 72.7\%), demonstrating its effectiveness in identifying out-of-distribution samples and mitigating their impact. When both modules are combined, the model achieves the best performance: accuracy of 65.0\%, AUC of 85.0\% and H score of 73.5\%, confirming their complementary and synergistic roles in enhancing generalization during testing.

\paragraph{OOD Detection Hyperparameter Analysis} 
\label{ood_hyper}
\begin{wrapfigure}[14]{r}{0.5\textwidth}
    \small
    \begin{center}
    \includegraphics[width=0.48\textwidth]{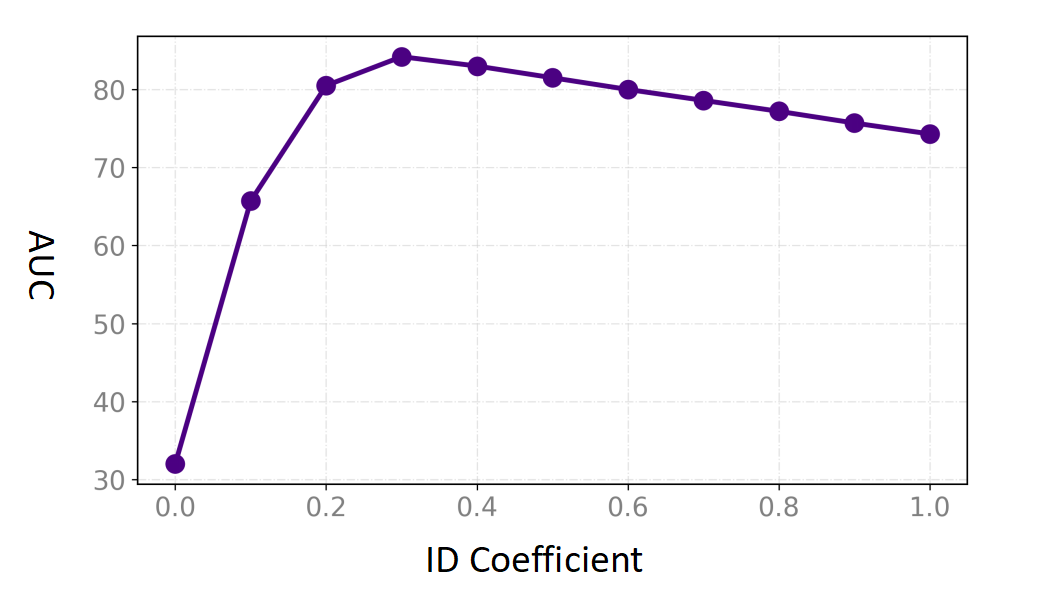}
    \end{center}
    \caption{Comparison of the impact of different ID coefficient hyperparameters on AUC scores.}
    \label{fig:label_based}
\end{wrapfigure}

In our experiments, we leverage auxiliary judgment information to generate enhanced OOD scores. Although maximizing entropy can help separate OOD samples from ID samples to some extent, our approach relies on adaptive fine-tuning with test data and does not have access to true labels, which prevents the imposition of explicit constraints on ID samples. Consequently, this limitation may indirectly increase the entropy of certain ID samples. To improve OOD detection while minimizing impact on the original classification performance, we introduce a balancing hyperparameter called the ID coefficient. Figure~\ref{fig:label_based} shows the distribution of this hyperparameter, where we adopt values of 0.3 and 0.7 in our experiments.

\section{Conclusion}

In this work, we propose a Hierarchical Ladder Network (HLN) that extracts OOD features by aggregating class tokens across all Transformer layers. OOD detection is improved by fusing the original model predictions with the HLN outputs through weighted probability fusion. Additionally, we introduce an Attention Affine Network (AAN) that adaptively refines the self-attention mechanism based on token information, enabling better adaptation to domain shifts and thereby enhancing classification performance on datasets with domain drift. Furthermore, a weighted entropy mechanism is employed to dynamically suppress the impact of low-confidence samples during adaptation. Experimental results on benchmark datasets demonstrate that our method significantly boosts performance on widely used classification tasks.

\section*{Acknowledgments}
This work was supported by the National Natural Science Foundation of China (Grant No. 12426312), Project 2021JC02X103, and the Center for Computational Science and Engineering at the Southern University of Science and Technology.

\bibliographystyle{plain}
\bibliography{refer}

\newpage
\appendix

\section{Details of Experimental Setup}
\subsection{Baseline and Backbone}
In this study, most baseline implementations and their corresponding training strategies were adapted from the official STAMP codebase \cite{yu2024stamp} to ensure faithful reproduction and fair comparison across methods. Although our overall procedure closely follows the original STAMP implementation, we adopt a Transformer-based backbone architecture and re-tune several hyperparameters—particularly the learning rate, weight decay, and adaptation-specific configurations—to better align with the characteristics of Transformer models. This design ensures both fairness and effectiveness within a unified experimental framework. Our choice is further motivated by empirical evidence suggesting that Vision Transformers achieve superior performance on large-scale datasets such as ImageNet. Experimental results corroborate this observation, showing that across most methods, the Vision Transformer backbone consistently outperforms the ResNet backbone used in STAMP in terms of both accuracy and robustness.

\subsection{Hyperparameters}
For all compared methods, including our proposed approach, hyperparameters are tuned consistently using the H-score metric on the Textures dataset \cite{cimpoi2014describing}, which serves as an out-of-distribution (OOD) benchmark. Specifically, we select the hyperparameter configuration that yields the best performance on this dataset and apply it directly to evaluations across all other target domains and outlier categories. This strategy ensures both consistency and fairness throughout all evaluations. The final hyperparameter settings for each method are summarized as follows:

\noindent \textbf{Ours.} \quad The procedure is outlined in Algorithm~\ref{alg:unified_tta}. We set the learning rates (LR) for the main network to 0.01, 0.02, 0.05, and 0.005 for ImageNet-C, ImageNet-R, ImageNet-A, and VisDA-2021, respectively. The learning rates for the QKV affine network are set to 0.0005, 0.0005, 0.0001, and 0.0005 in the same order. For the class token feature extraction network $\Psi$, we use learning rates of 0.1, 0.1, 0.1, and 0.01. The learning rates for the Hierarchical Ladder Network are configured as 0.001, 0.001, 0.0001, and 0.001, respectively.

\noindent \textbf{Tent.} \quad For Tent~\cite{wang2021tent}, we set the learning rate to $\text{LR} = 0.005$ for ImageNet-C. We followed the official implementation\footnote{\url{https://github.com/DequanWang/tent}}.

\noindent \textbf{CoTTA.} \quad For CoTTA~\cite{wang2022continual}, we set the restoration factor to $p = 0.0005$ and the augmentation confidence threshold to $p_{\text{th}} = 0.05$. We followed the official implementation\footnote{\url{https://github.com/qinenergy/cotta}}.

\noindent \textbf{SoTTA.} \quad For SoTTA~\cite{gong2023sotta}, we set the learning rate to $\text{LR} = 0.005$ and the confidence threshold to $C_0 = 0.33$. Other hyperparameters follow the original paper. We followed the official implementation\footnote{\url{https://github.com/taeckyung/SoTTA}}.

\noindent \textbf{SAR.} \quad For SAR~\cite{niu2023towards}, the learning rates are set to 0.01, 0.02, 0.05, and 0.005 for ImageNet-C, ImageNet-R, ImageNet-A, and VisDA-2021, respectively. The reset threshold is fixed at 0.1. We followed the official implementation\footnote{\url{https://github.com/mr-eggplant/SAR}}.

\noindent \textbf{OSTTA.} \quad For OSTTA~\cite{lee2023towards}, we set the learning rate to 0.001 on ImageNet-C and the loss coefficient $\lambda_{1}$ to 0.2. Since the official implementation is not available, we adopt the re-implementation from UniEnt\footnote{\url{https://github.com/gaozhengqing/UniEnt}}.

\noindent \textbf{UniEnt.} \quad For UniEnt~\cite{niu2023towards}, we set the learning rate to 0.001 on ImageNet-C. The hyperparameters $\lambda_{1}$ and $\lambda_{2}$ are set to 0.1 and 0.1, respectively. We followed the official implementation\footnote{\url{https://github.com/gaozhengqing/UniEnt}}.

\noindent \textbf{STAMP.} \quad For STAMP~\cite{yu2024stamp}, the learning rates are set to 0.05, 0.1, 0.1, and 0.1 for ImageNet-C, ImageNet-R, ImageNet-A, and VisDA-2021, respectively. The threshold scaling factor $\alpha$ is fixed at 0.8. We followed the official implementation\footnote{\url{https://github.com/yuyongcan/STAMP}}.

\subsection{Hyperparameters for ViT-L}
For this evaluation, the learning rate for the normalization layers is set to 0.0001. For the QKV Affine Network, $\Psi$, and the Hierarchical Ladder Network, the learning rates are configured as 0.00005, 0.0001, and 0.001, respectively. The reset constant and the learning rate for LR are set to 0.0001 and 0.2, respectively. For STAMP, the learning rate and the scaling factor $\alpha$ are set to 0.01 and 0.8. The batch size is fixed at 16.

\begin{algorithm}[!t]
\caption{Unified Test-Time Adaptation with HLN and AAN under Open-World Setting}
\label{alg:unified_tta}

\KwIn{Source pre-trained model $f_{\theta_s}$; target stream $\{X_t\}$}
\KwOut{Predictions $\{\hat{\mathbf{y}}\}$ on target data}

Initialize target model $f_{\theta_t} \leftarrow f_{\theta_s}$\;
Initialize optimizer with tunable parameters $\tilde{\theta}$ and learning rate $\eta$\;

\ForEach{batch $\mathbf{B}_j \subset \{X_t\}$}{
    \textbf{Step 1: AAN-based Feature Adaptation} \\
    Extract patch tokens $E$\;
    Apply Attention Affine Network (AAN) to adapt QKV projections using $E$\;
    
    \textbf{Step 2: HLN-based OOD Modeling} \\
    Extract class tokens $\mathbf{c}_{\text{cls}}^{(l)}$ from each transformer layer\;
    Compute OOD tokens: $\mathbf{o}^{(l)} = \Psi^l(\mathbf{c}_{\text{cls}}^{(l)})$\;
    Aggregate via HLN: $\mathbf{o}^{\text{hln}} = \mathrm{HLN}(\{ \mathbf{o}^{(l)} \}_{l=1}^{L})$\;
    
    \textbf{Step 3: Forward Pass and Loss Computation} \\
    Compute OOD predictions: \\
    $\mathbf{p}^{\text{final}} = \alpha \cdot \mathrm{softmax}(\mathcal{C}(\mathbf{c}_{\text{cls}}^{(L)})) + (1 - \alpha) \cdot \mathrm{softmax}(\mathcal{C}(\mathbf{o}^{\text{hln}}))$\;
    Compute entropy $\mathcal{H}_i$ for each sample\;
    Compute self-weighted entropy loss $\mathcal{L}_{\text{entropy}}$\;
    Compute patch similarity loss $\mathcal{L}_{\text{sim}}$ based on cosine similarity among patch tokens\;
    Compute OOD loss $\mathcal{L}_{\text{OOD}}$ using masked high-entropy samples\;
    
    \textbf{Step 4: Model Update and Prediction} \\
    Compute total loss: \\
    $\mathcal{L} = \mathcal{L}_{\text{entropy}} + \alpha \mathcal{L}_{\text{OOD}} + \beta \mathcal{L}_{\text{sim}}$\;
    Update model parameters: $\tilde{\theta} \leftarrow \tilde{\theta} - \eta \cdot \nabla \mathcal{L}$\;
    Output predictions: $\hat{\mathbf{y}} = f_{\theta_t}(\mathbf{B}_j)$\;
}
\end{algorithm}

\section{Limitations and future work}
Although we have evaluated the overall classification performance of the model under input distribution shifts and image quality degradation by testing on multiple diverse datasets—thereby simulating various scenarios where the model encounters samples different from the training set—the real-world testing environment is often more complex. It may involve a wider variety of data types, more diverse out-of-distribution (OOD) interferences, and more challenging analytical scenarios such as those encountered in 3D settings. This work does not fully address such diversity and complexity. Recent studies \cite{jiang2024pcotta} have explored the model’s adaptability to continuously evolving target domains in 3D environments. In the future, maintaining stable model performance when confronted with novel OOD category samples in these more complex scenarios remains a practical and significant challenge.

\section{Supplementary experiments}

\begin{table*}[!htbp]

\begin{center}
\caption{Classification Accuracy (\%) for each corruption in \textbf{Visda-2021} at the highest severity (Level 5). The best result is shown in \textbf{bold}.}
\label{table:visda}
\resizebox{\linewidth}{!}
{
\begin{tabular}{l|ccc|ccc|ccc|ccc|ccc}
\toprule
\multirow{2}{*}{Source} & \multicolumn{3}{c|}{Textures} & \multicolumn{3}{c|}{Places} & \multicolumn{3}{c|}{iNaturalist} & \multicolumn{3}{c|}{SUN} & \multicolumn{3}{c}{Avg.} \\
& ACC & AUC & H-score & ACC & AUC & H-score & ACC & AUC & H-score & ACC & AUC & H-score & ACC & AUC & H-score \\
	
\midrule
Source & 44.3 & 59.0 & 50.6 & 44.3 & 51.1 & 47.4 & 
44.3 & 72.3 & 54.9 & 44.3 & 56.5 & 49.7 & 44.3 & 59.7 & 50.6 \\ 
SAR & 50.7 & 63.3 & 56.3 & 50.5 & 55.8 & 53.1 & 
51.0 & 78.7 & 61.9 & 50.6 & 61.7 & 55.4 & 50.7 & 64.9 & 56.6 \\ 
STAMP & 57.2 & 66.9 & 61.7 & 57.4 & 66.1 & 61.4 & 
57.1 & 83.8 & 67.9 & 57.1 & 59.1 & 58.1 & 57.2 & 69.0 & 62.3 \\ 
\rowcolor{gray!20}
\textbf{Ours} & \textbf{58.5} & \textbf{70.8} & \textbf{64.1} & \textbf{58.6} & \textbf{56.4} & \textbf{57.5} & 
\textbf{58.7} & \textbf{89.9} & \textbf{71.0} & 
\textbf{57.9} & \textbf{70.5} & \textbf{63.6} & 
\textbf{58.5} & \textbf{71.9} & \textbf{64.0} \\ 
\rowcolor{gray!20}
& ${\pm0.5}$ & ${\pm0.5}$ & ${\pm0.3}$ & 
${\pm0.4}$ & ${\pm1.1}$ & ${\pm0.8}$ & 
${\pm0.7}$ & ${\pm0.2}$ & ${\pm0.6}$ & 
${\pm0.5}$ & ${\pm1.0}$ & ${\pm0.6}$ & 
${\pm0.5}$ & ${\pm0.7}$ & ${\pm0.6}$ \\ 
\bottomrule
\end{tabular}
}
\end{center}
\end{table*}

\subsection{Additional Results on VisDA-2021}

To comprehensively evaluate the cross-domain adaptability of the models, we conducted experiments on the \textbf{VisDA-2021} dataset~\cite{bashkirova2022visda}. This dataset contains approximately 20,000 images from both synthetic and real domains, covering 12 object categories, and is widely regarded as a standard benchmark for domain adaptation research. To further assess out-of-distribution (OOD) robustness, we additionally incorporated \textbf{Textures}~\cite{cimpoi2014describing}, \textbf{Places}~\cite{zhou2017places}, \textbf{iNaturalist}~\cite{van2018inaturalist}, and \textbf{SUN}~\cite{xiao2010sun} as potential OOD categories during testing. The corresponding results are summarized in Table~\ref{table:visda}.

\subsection{Ablation Study on the Hyperparameters}

\textbf{Sharpness-Aware Optimization with Custom Loss.} To enhance robustness against domain shift and open-set samples during test time, we adopt the Sharpness-Aware Minimization (SAM) strategy~\cite{foret2020sharpness}, which encourages the model to converge to flat minima by optimizing over the worst-case local neighborhood. In our implementation, we define two customized loss functions used for SAM's two-step optimization:

In the first backward pass, the loss is formulated as:
\begin{equation}
\mathcal{L}_1(x; \Theta) = \mathcal{L}_{\text{entropy}}(x; \Theta) + \lambda_1 \cdot \mathcal{L}_{\text{OOD}}(x; \Theta) + \mathcal{L}_{\text{sim}}(x; \Theta),
\label{eq:loss1}
\end{equation}
where $\mathcal{L}_{\text{entropy}}$ denotes the entropy loss, $\mathcal{L}_{\text{OOD}}$ is the out-of-distribution (OOD) token loss, and $\mathcal{L}_{\text{sim}}$ represents the inter-class similarity loss.

Next, we compute the adversarial perturbation in parameter space via a first-order approximation:
\begin{equation}
\hat{\epsilon}(\Theta) = \rho \cdot \frac{\nabla_\Theta \mathcal{L}_1(x; \Theta)}{\left\| \nabla_\Theta \mathcal{L}_1(x; \Theta) \right\|_2},
\label{eq:eps}
\end{equation}
and perform a virtual ascent step to obtain the perturbed weights $\Theta + \hat{\epsilon}$.

In the second backward pass, the loss is defined as:
\begin{equation}
\mathcal{L}_2(x; \Theta + \hat{\epsilon}(\Theta)) = \mathcal{L}_{\text{entropy}}(x; \Theta + \hat{\epsilon}(\Theta)) + \lambda_2 \cdot \mathcal{L}_{\text{OOD}}(x; \Theta + \hat{\epsilon}(\Theta)),
\label{eq:loss2}
\end{equation}
which is then used to compute the final gradient for parameter updates.

\begin{figure}[htbp]
  \centering
  \includegraphics[width=0.8\linewidth]{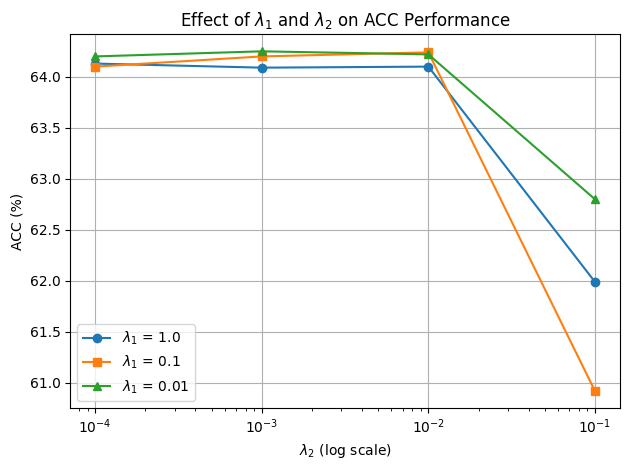}
  \caption{The influence of regularization parameters $\lambda_1$ and $\lambda_2$ on model accuracy (ACC). The curves reflect the trend of accuracy changes with varying parameter values.}
  \label{fig:acc}
\end{figure}

\begin{figure}[htbp]
  \centering
  \includegraphics[width=0.8\linewidth]{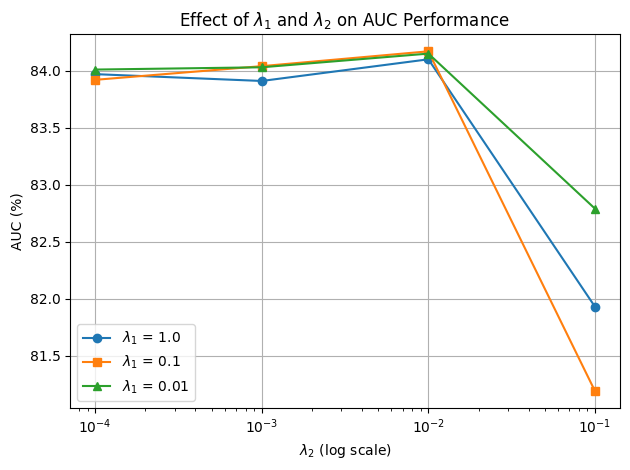}
  \caption{The influence of regularization parameters $\lambda_1$ and $\lambda_2$ on auc score (AUC). The curves reflect the trend of accuracy changes with varying parameter values.}
  \label{fig:auc}
\end{figure}

\begin{figure}[htbp]
  \centering
  \includegraphics[width=0.8\linewidth]{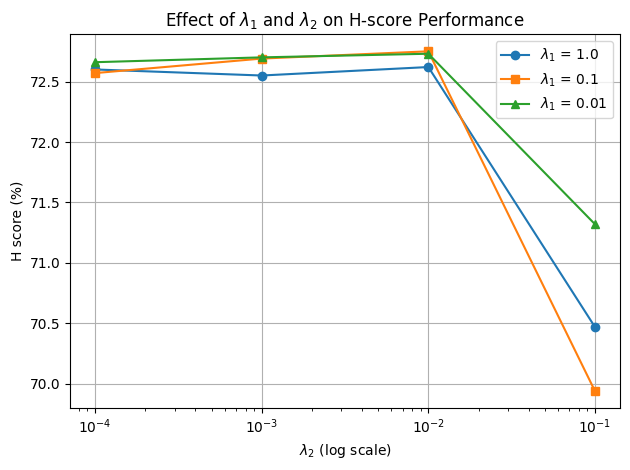}
  \caption{The influence of regularization parameters $\lambda_1$ and $\lambda_2$ on h score (H-score). The curves reflect the trend of accuracy changes with varying parameter values.}
  \label{fig:hscore}
\end{figure}

This two-step optimization enhances both feature discriminability and parameter smoothness, thereby improving the model’s robustness to distributional shifts. Based on comparisons of accuracy (Figure~\ref{fig:acc}), AUC (Figure~\ref{fig:auc}), and H-score (Figure~\ref{fig:hscore}), we select \(\lambda_1 = 0.01\) and \(\lambda_2 = 0.001\) as the hyperparameters used in our experiments.

\begin{table}[htbp]
\normalsize
\centering
\caption{Running time comparison (in minutes) on an NVIDIA RTX 3090 GPU. 
All methods are evaluated with ViT-B on ImageNet-C (Gaussian noise, severity level 5), 
using Textures as the OOD dataset (55,000 images).}
\label{tab:running_time_horizontal}
\resizebox{\linewidth}{!}{
\begin{tabular}{lccccccccccc}
\toprule
Method & Source  & TENT  & CoTTA & SoTTA & OSTTA & UniEnt & SAR & STAMP & HLN & AAN & Ours \\
\midrule
Time   & $\sim$3  & $\sim$6 & $\sim$79 & $\sim$22 & $\sim$8 & $\sim$11 & $\sim$10 & $\sim$75  & $\sim$10 & $\sim$25 & $\sim$26 \\
\bottomrule
\end{tabular}}
\end{table}

\subsection{Running Time Analysis}

To provide a more comprehensive understanding of the computational efficiency of different TTA methods, 
we report the overall running time for completing the evaluation on ViT-B with ImageNet-C (Gaussian noise, severity level 5) 
and Textures as the OOD dataset (55,000 images). 
Table~\ref{tab:running_time_horizontal} summarizes the comparison results. 
As shown, lightweight approaches such as Source Only and OSTTA finish within several minutes, 
while methods involving iterative updates or auxiliary modules (e.g., STAMP, CoTTA, UniEnt, HLN, and AAN) require significantly longer time. 
Our combined method (HLN + AAN) achieves a balanced trade-off between efficiency and performance, 
finishing in around half an hour.

\end{document}